\documentclass[letterpaper, 10 pt, conference]{ieeeconf}  % Comment this line out if you need a4paper

\IEEEoverridecommandlockouts                              % This command is only needed if 
\usepackage{graphicx} % for pdf, bitmapped graphics files
\usepackage{amsmath} % assumes amsmath package installed
\usepackage{amssymb}  % assumes amsmath package installed
\usepackage{gensymb}
\usepackage{booktabs}
\usepackage{multirow}
\usepackage{xcolor}
\usepackage[
    bookmarks=true,
    colorlinks=true,  % Enables link coloring
    linkcolor=red,    % Color for internal links
    citecolor=green,  % Color for citations (references)
    filecolor=magenta, % Color for file links
    urlcolor=orange,     % Color for external links
    anchorcolor=blue
]{hyperref}

\title{\LARGE \bf
NeuGrasp: Generalizable Neural Surface Reconstruction with Background Priors for Material-Agnostic Object Grasp Detection
}

\author{Qingyu Fan$^{1,2,3}$, Yinghao Cai$^{1,2\dag}$, Chao Li$^{3}$, Wenzhe He$^{3}$, Xudong Zheng$^{3}$, Tao Lu$^{1}$, Bin Liang$^{3}$, Shuo Wang$^{1, 2}$% <-this % stops a space
\thanks{$^{1}$State Key Laboratory of Multimodal Artificial Intelligence Systems, Institute of Automation, Chinese Academy of Sciences.}%
\thanks{$^{2}$School of Artificial Intelligence, University of Chinese Academy of Sciences.}%
\thanks{$^{3}$Qiyuan Lab.}%
\thanks{*This work was supported in part by the National Natural Science Foundation of China under Grants 62273342, 62473366 and U23B2038, and in part by the Qiyuan Lab Innovation Fund Project (2022-JCJQ-LA-001-024).}
\thanks{$\dag$Corresponding to {\tt \small yinghao.cai@ia.ac.cn}.}%
}

\begin{document}

\maketitle
\thispagestyle{empty}
\pagestyle{empty}

%%%%%%%%%%%%%%%%%%%%%%%%%%%%%%%%%%%%%%%%%%%%%%%%%%%%%%%%%%%%%%%%%%%%%%%%%%%%%%%%
\begin{abstract}

%=============Qingyu==========Robotic grasping in cluttered environments with diverse materials, including transparent and specular surfaces, poses challenges for conventional depth-sensing methods. We introduce NeuGrasp, a neural surface reconstruction method that leverages background priors for material-agnostic grasp detection. NeuGrasp integrates transformers and global prior volumes to aggregate multi-view features with spatial encoding, enabling robust surface reconstruction even in highly narrow and sparse viewing conditions. Our innovative use of background priors enhances focus on foreground objects via residual feature enhancement and refines spatial perception with an occupancy-prior volume, particularly for transparent and specular objects. Extensive experiments in both simulated and real-world settings show NeuGrasp significantly outperforms state-of-the-art methods in grasping while maintaining comparable reconstruction quality. Moreover, NeuGrasp-RA (Reality Augmentation), a fine-tuned variant with small-scale real-world data, demonstrates strong domain adaptation, proving its robustness in practical scenarios. 

Robotic grasping in scenes with transparent and specular objects presents great challenges for methods relying on accurate depth information.
In this paper, we introduce NeuGrasp, a neural surface reconstruction method that leverages background priors for material-agnostic grasp detection. NeuGrasp integrates transformers and global prior volumes to aggregate multi-view features with spatial encoding, enabling robust surface reconstruction in narrow and sparse viewing conditions. By focusing on foreground objects through residual feature enhancement and refining spatial perception with an occupancy-prior volume, NeuGrasp excels in handling objects with transparent and specular surfaces. Extensive experiments in both simulated and real-world scenarios show that NeuGrasp outperforms state-of-the-art methods in grasping while maintaining comparable reconstruction quality. More details are available at \url{https://neugrasp.github.io/}.

\end{abstract}

%%%%%%%%%%%%%%%%%%%%%%%%%%%%%%%%%%%%%%%%%%%%%%%%%%%%%%%%%%%%%%%%%%%%%%%%%%%%%%%%
\section{INTRODUCTION}
Vision based 6-DoF grasping in robotics remains a significant challenge due to the wide diversity of object shapes, textures, and material properties in real-world environments \cite{fang2020graspnet}, \cite{sundermeyer2021contact}, \cite{breyer2021volumetric}, \cite{zheng2022vgpn}, \cite{zheng2023gpdan}, \cite{fang2023anygrasp}. Accurate reconstruction of 3D scene geometry  is essential for determining feasible grasps.  However, this task is often compromised by limitations of depth sensors which produce unreliable depth measurements when confronted with transparent and specular objects. Such limitations  result in degraded geometric representations of the scene and, consequently, unsuccessful grasps.

%======Qingyu========Object grasping in robotics remains an enduring challenge, primarily due to the diverse spectrum of shapes, textures, and material compositions encountered in real-world environments. Accurate 3D geometry reconstruction is essential for determining feasible grasping poses; however, this task is often compromised by the inherent limitations of depth sensors. Although these sensors have been integral to earlier approaches \cite{fang2020graspnet}, \cite{sundermeyer2021contact}, \cite{breyer2021volumetric}, \cite{zheng2022vgpn}, \cite{zheng2023gpdan}, \cite{fang2023anygrasp} for capturing scene geometry, they falter when confronted with transparent and specular objects, which are ubiquitous across daily settings, industrial applications, and research laboratories. This often results in unreliable depth data, culminating in deteriorated geometric representations and, consequently, unsuccessful grasping attempts.

%===============Qingyu    \caption{\textbf{Overview of NeuGrasp.} We introduce a generalizable method that leverages background priors within a neural implicit surface reconstruction framework to achieve real-time, depth-independent scene reconstruction and grasp detection, effectively handling sparse views, narrow fields of view, and challenging object materials.}

\begin{figure}[t]
    \centering
    \includegraphics[width=0.48\textwidth]{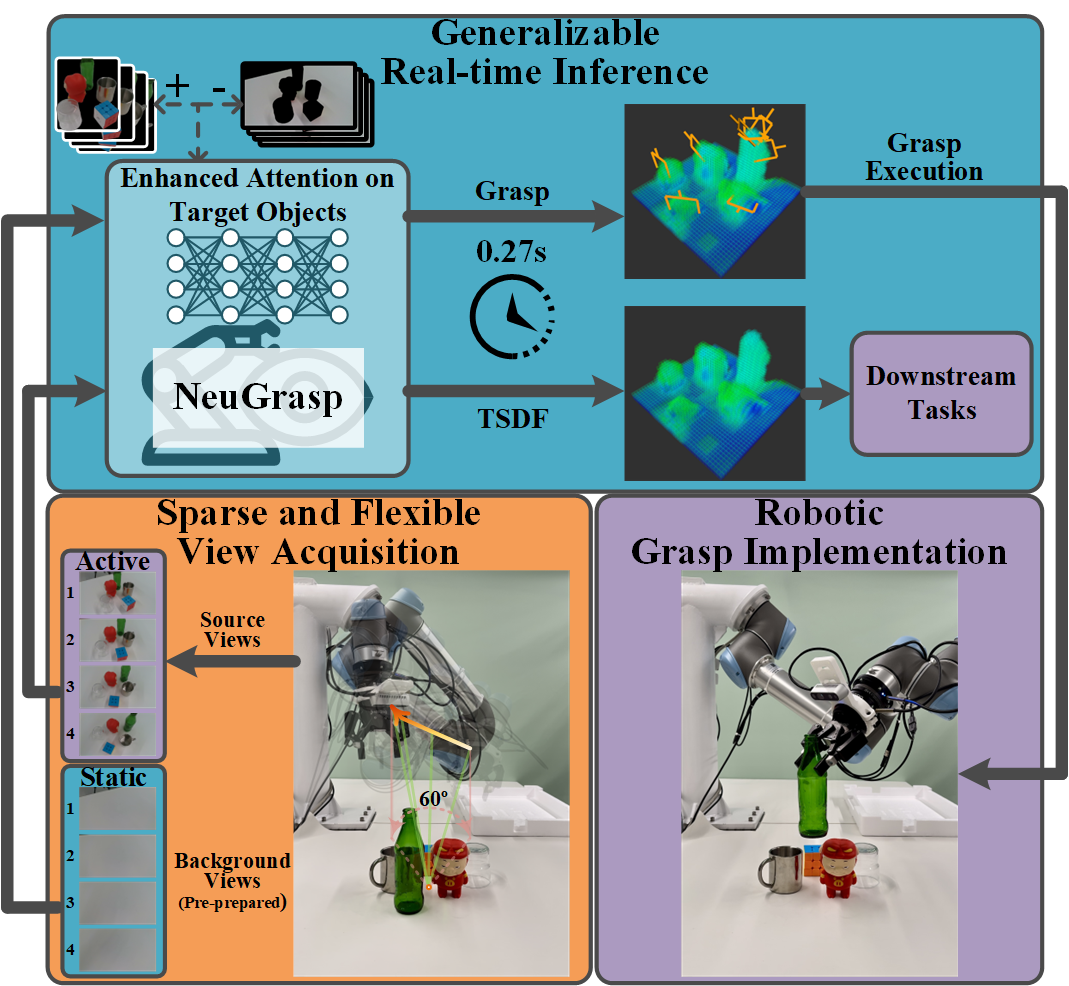}
    \caption{\textbf{Overview of NeuGrasp.} 
    We introduce a generalizable method that utilizes background priors within a neural implicit surface framework to achieve real-time scene reconstruction and material-agnostic grasping from observations within a narrow field of view.}
    \vspace{-0.5cm}
    \label{fig1}
  
\end{figure}

Recent research has explored the use of neural radiance field (NeRF) \cite{mildenhall2020nerf} to address the challenge of grasping transparent and specular objects due to its ability to effectively capture complex light interactions on non-Lambertian surfaces. DexNeRF \cite{ichnowski2022dex} pioneers this approach but it suffers from the need for dense input images and extensive per-grasp training, making it impractical for real-world applications. GraspNeRF \cite{dai2023graspnerf} uses a generalizable NeRF that requires only sparse views and  eliminates the need for per-scene optimization. However, it still relies on  comprehensive 360-degree image capture and ground-truth Truncated Signed Distance Function (TSDF) supervision, which may not always be feasible in real-world scenarios. RGBGrasp \cite{liu2024rgbgrasp} further improves upon this by integrating an all-in-one NeRF extension from \cite{tancik2023nerfstudio} with pre-trained models for depth estimation and grasp detection, achieving robust performance with RGB-only inputs. However, it still relies on dense views and lacks generalizability, requiring time-consuming retraining for each grasp, which limits its use for real-time applications.

%===============Qingyu==========Recent research has leveraged the neural radiance field (NeRF) \cite{mildenhall2020nerf} to tackle the challenges of grasping transparent and specular objects due to its ability to effectively capture complex light interactions on non-Lambertian surfaces. DexNeRF \cite{ichnowski2022dex} pioneered this approach but is hampered by its reliance on dense input images and extensive per-grasp training, making it impractical for real-world use and limiting it to 3-DoF grasps. GraspNeRF \cite{dai2023graspnerf} improves practicality by using a generalizable NeRF that requires only sparse views, eliminating per-scene optimization. However, it still demands 360-degree image capture and ground-truth TSDF supervision, restricting its effectiveness in real-world scenarios with limited viewpoints and no available depth maps. RGBGrasp \cite{liu2024rgbgrasp} enhances this by integrating an all-in-one NeRF extension from \cite{tancik2023nerfstudio} with pre-trained models for depth estimation and grasp perception, achieving robust performance with RGB-only inputs. Yet, it still relies on dense views and lacks generalizability, requiring time-consuming retraining after each grasp, making it unsuitable for real-time applications.

In this paper, we introduce \textbf{NeuGrasp} for the 6-DoF robotic grasping tasks (Fig. \ref{fig1}). NeuGrasp is a generalizable method that efficiently utilizes background priors to reconstruct scene geometry and generate diverse grasp candidates from observations within a narrow field of view, all in real time and without the need for depth-related supervision. By employing Transformer architecture within a neural implicit reconstruction framework, NeuGrasp combines multi-view features enriched with spatial encoding to infer scene geometry, thereby facilitating surface reconstruction in novel grasping scenarios. 
Additionally, the integration of a high-level global prior volume allows NeuGrasp to perform effectively even under narrow fields of view and sparse viewing conditions. 

A key aspect of our approach is the use of background priors for surface reconstruction with transparent and specular objects. We propose a residual feature enhancement module which enhances the model’s attention on foreground objects by contrasting features from scene and background images. Additionally, we introduce an occupancy-prior volume that utilizes global implicit occupancy information from residual feature maps to enhance spatial perception, especially for transparent and specular objects. 

%================Qingyu-----------Experiments are carried out in both simulated and real-world robotic environments. Our results show that NeuGrasp achieves an average improvement in grasp success rate of 30.33\% in the pile setting and 41.86\% in the packed setting, significantly outperforming baselines in challenging scenarios that involve only transparent and specular objects. Remarkably, it also rivals the surface reconstruction performance of methods that directly utilize explicit geometry supervision. Moreover, without relying on depth-related supervision, additional real-world experiments demonstrate that \textbf{NeuGrasp-RA (Reality Augmentation)}, an enhanced version fine-tuned with small-scale real data, significantly boosts performance by effectively bridging the domain gap between simulation and reality. These findings underscore the effectiveness and exceptional capabilities of our approach. In summary, our contributions are:

Experimental results show that NeuGrasp significantly outperforms baselines especially in  scenarios that involve only transparent and specular objects. The performance of surface reconstruction of NeuGrasp is on par with the methods that use explicit geometry supervision. With finetuning on a small real-world dataset, 
NeuGrasp-RA (Reality Augmentation) further improves the performance of robotic grasping which demonstrates the potential of our approach to be applied in real applications. In summary, our contributions are:

%    \item We integrate a view transformer, a ray transformer, and a global prior volume to hierarchically aggregate multi-view features and spatial priors, providing a foundation for generalizable implicit surface reconstruction under narrow and sparse view conditions.\item We propose an effective strategy for leveraging scene background priors, utilizing residual features to enhance attention to target objects and improve spatial perception, empowering end-to-end reconstruction and grasping synergy under extremely unfavorable viewing conditions.\item We comprehensively demonstrate the superiority of our method over the state-of-the-art and, through fine-tuning with a small-scale dataset collected from the real world, obtain a highly performant model, thereby proving the benefits of not requiring depth-related supervision.

\begin{itemize}
% [itemsep=0pt, topsep=0pt]
    \item We integrate a view transformer,  ray transformer, and global prior volume to hierarchically aggregate multi-view features and spatial priors, providing an effective solution for generalizable implicit surface reconstruction under narrow and sparse view conditions. Both surface reconstruction and grasp detection are trained in an end-to-end manner.
    \item We propose to leverage scene background priors for surface reconstruction in scenes with transparent and specular objects. Through residual feature enhancement, these objects can be clearly distinguished from the background, facilitating more accurate grasp detection. 
    \item We demonstrate the superiority of our method in both simulated and real-world experiments for the tasks of reconstruction and grasp detection. NeuGrasp significantly outperforms baselines in grasping particularly in scenarios with transparent and specular objects.
\end{itemize}

\section{RELATED WORK}

\subsection{Generalizable Neural Radiance Field}

NeRF \cite{mildenhall2020nerf}, initially proposed for novel view synthesis using volume rendering and implicit neural representations, has since been extended to scene reconstruction and augmented reality. Some existing works \cite{tancik2023nerfstudio}, \cite{barron2021mip}, \cite{barron2022mip}, \cite{muller2022instant}, \cite{wang2023sparsenerf} rely on per-scene optimization, which limits the generalization to unseen scenes. IBRNet \cite{wang2021ibrnet} introduces generalization by blending nearby views using MLPs and ray transformers. NeuRay \cite{liu2022neural} considers the view visibility. ContraNeRF \cite{yang2023contranerf} introduces contrastive learning for synthetic-to-real generalization.

%======================Qingyu=============NeRF \cite{mildenhall2020nerf}, initially developed for novel view synthesis through volume rendering and implicit neural representations, has since expanded to scene reconstruction and augmented reality. However, many follow-up works \cite{zhang2020nerf++}, \cite{barron2021mip}, \cite{muller2022instant}, \cite{wang2023sparsenerf}, \cite{tancik2023nerfstudio} rely on per-scene optimization, limiting their generalization to unseen scenes. IBRNet \cite{wang2021ibrnet} improves generalization by blending nearby views using MLPs and ray transformers, while NeuRay \cite{liu2022neural} considers view visibility. ContraNeRF \cite{yang2023contranerf} introduces contrastive learning for synthetic-to-real generalization.

%==========================Qingyu ===========NeRF's implicit field is less suited for spatial tasks like grasp detection, often resulting in flawed reconstructions. NeuS \cite{wang2021neus} addresses this with signed distance functions (SDFs) for more accurate multi-view reconstruction. SparseNeuS \cite{long2022sparseneus} uses hierarchical feature volumes, and C2F2NeuS \cite{xu2023c2f2neus} integrates MVS with implicit reconstruction. VolRecon \cite{ren2023volrecon} and ReTR \cite{liang2024retr} use transformers for multi-view feature fusion, refining details and surface accuracy, while UFORecon \cite{na2024uforecon} enhances robustness with cross-view matching transformers for challenging viewing conditions.

However, the implicit field representation of NeRF may produce inaccurate surface reconstruction for the task of grasp detection. NeuS \cite{wang2021neus} addresses this issue using the signed distance function (SDF) for more accurate multi-view reconstructions. Other methods such as SparseNeuS \cite{long2022sparseneus} use hierarchical feature volumes, and C2F2NeuS \cite{xu2023c2f2neus} combines multi-view stereo (MVS) with implicit reconstruction. VolRecon \cite{ren2023volrecon} and ReTR \cite{liang2024retr} use transformer for multi-view feature fusion, which helps refine details and improves accuracy of surface reconstruction. Furthermore, UFORecon \cite{na2024uforecon} improves the robustness of neural surface with cross-view matching transformers in challenging viewing conditions.

\begin{figure*}[t]
    \centering
    \includegraphics[width=\textwidth]{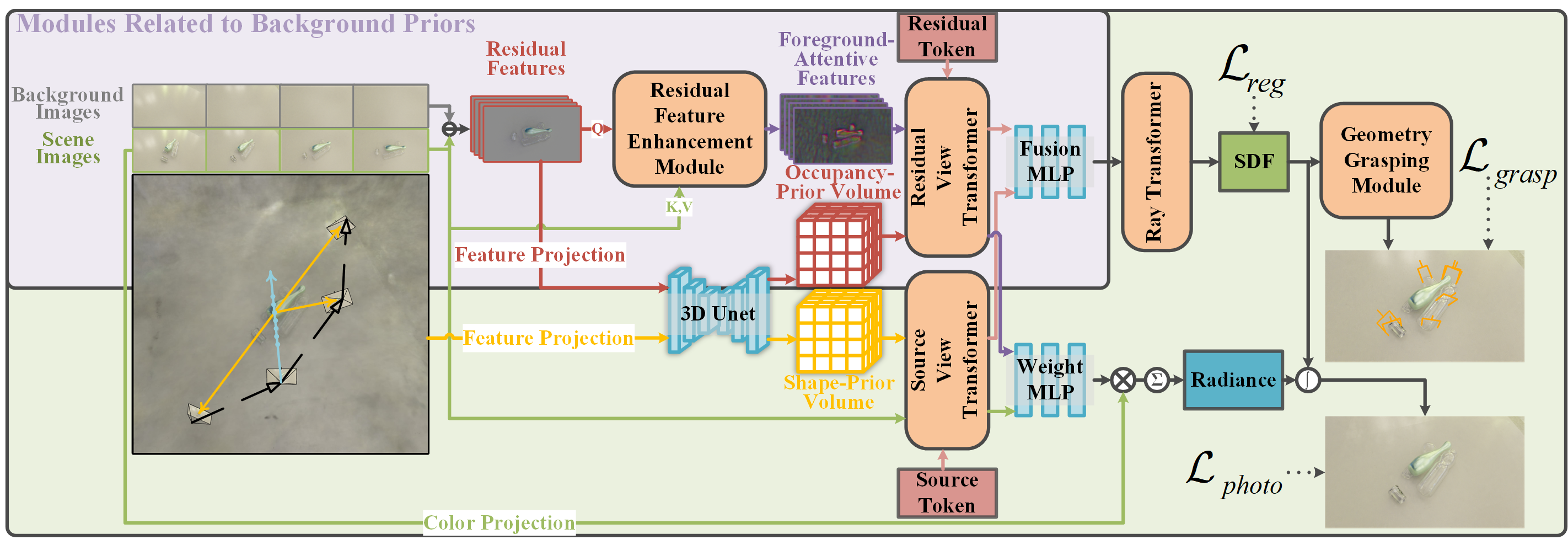}
    \caption{\textbf{Framework of NeuGrasp.} NeuGrasp leverages background priors for  
    neural surface reconstruction and material-agnostic grasp detection. A Residual Feature Enhancement module is proposed to enhance the model attention on foreground objects instead of irrelevant background information. 
    Through feature projection, we construct an occupancy-prior volume from residual features and a shape-prior volume from scene features. 
    These volumes are then separately combined with their corresponding multi-view features using View Transformers. After fusion, a Ray Transformer further refines the spatial information. The final reconstructed geometry is represented as a signed distance function and converted into a radiance field.
    Finally, the grasping module  maps the reconstructed geometry to 6-DoF grasp poses, enabling end-to-end training. }
    % through differentiable volumetric rendering. }
    \label{fig2}
    \vspace{-0.5cm}
\end{figure*}

\subsection{NeRF in Robotic Grasping}

Grasp pose detection using RGB-D sensors has been extensively studied for decades \cite{fang2020graspnet}, \cite{sundermeyer2021contact}, \cite{breyer2021volumetric}, \cite{zheng2022vgpn}, \cite{zheng2023gpdan}, \cite{fang2023anygrasp}. However, grasp pose detection from point clouds continues to face challenges with transparent and specular objects due to inaccuracies in depth maps. NeRF provides a solution to address this issue by modeling light propagation, as seen in DexNeRF \cite{ichnowski2022dex}. DexNeRF's reliance on dense inputs and extensive training requirements limit its practicality in real-world applications. EvoNeRF \cite{kerr2023evo} and MIRA \cite{lin2023mira} optimize NeRF for 6-DoF grasping but still require dense inputs. Residual-NeRF improves depth perception using static, opaque regions as priors but is impractical due to the limited availability of such priors in real scenarios. GraspNeRF \cite{dai2023graspnerf} removes the need for per-scene optimization, but requires 360-degree captures and ground-truth TSDF for supervision, which limits real-world applications. Similarly, 
RGBGrasp \cite{liu2024rgbgrasp} integrates an all-in-one NeRF extension but still requires dense inputs and retraining for each grasp. In contrast, we propose NeuGrasp, which overcomes these limitations by leveraging background priors and generalizable implicit surface reconstruction. NeuGrasp enables material-agnostic grasping in a narrow field of view without the need for geometric supervision.

%=============================Qingyu================Deep learning methods \cite{fang2020graspnet}, \cite{sundermeyer2021contact}, \cite{breyer2021volumetric}, \cite{zheng2022vgpn}, \cite{zheng2023gpdan}, \cite{fang2023anygrasp} have advanced cluttered grasping, but struggle with transparent and specular objects due to RGB-D depth map inaccuracies. NeRF offers a solution by modeling light propagation, as seen in DexNeRF \cite{ichnowski2022dex}, though its reliance on dense inputs and extensive training limits practicality. EvoNeRF \cite{kerr2023evo} and MIRA \cite{lin2023mira} optimized NeRF for 6-DoF control but still required dense inputs. Residual-NeRF \cite{duisterhof2024residual} improved depth perception but relied on strong scene priors. GraspNeRF \cite{dai2023graspnerf} eliminated per-scene optimization but required 360-degree captures and ground-truth TSDF, limiting real-world use. RGBGrasp \cite{liu2024rgbgrasp} integrated an all-in-one NeRF extension but still needed dense views and retraining, hindering real-time application. In contrast, NeuGrasp overcomes these limitations by using background priors and generalizable implicit surface reconstruction, enabling material-agnostic grasping in a narrow field of view without geometric supervision.
\section{METHOD}

\subsection{Problem Description}

% 给定沿狭窄视角范围沿轨迹拍摄的N张杂乱桌面静态场景图片，目标是重建场景的几何信息，并通过重建后的信息预测6自由度无碰撞抓取位姿，以完成桌面上物体的抓取操作。重建的场景信息会直接影响抓取的预测性能，模糊的重建场景信息会导致碰撞和不可达的抓取位置。然而，仅根据狭窄视角范围内的稀疏视图还原完整的场景信息是一个病态问题。我们假设机器人执行抓取的工作场景使静态固定的，基于此，我们通过引入背景先验信息，以实现少数据监督（仅RGB图片）的端到端重建-抓取预测。
% 如Fig2所示，输入N张场景图片和相应位置预先准备的背景图片，相机内参外参已知，NeuGrasp首先基于可泛化隐式神经重建的框架，引入transformer结构进行视角融合和几何预测，并通过残差特征增强和代价体积融合背景先验信息，重建出完整的场景TSDF表示。之后利用抓取检测网络，将预测的TSDF映射为一系列的6-DoF抓取预测。因为体积渲染架构的可微性，我们可以使用端到端的训练方式实现unseen场景的泛化，快速推理重建场景并生成无碰撞的6-DoF抓取。

%==============Qingyu========Given a sequence of images $\left\{ {{\mathbf{I}_i}} \right\}_{i = 1}^N$ from a fixed trajectory with a narrow field of view, the task is to reconstruct scene geometry and predict collision-free 6-DoF grasps. Sparse, limited views make this inherently ill-posed, but NeuGrasp addresses it by leveraging background priors for end-to-end reconstruction and grasp prediction without explicit geometry supervision.

Given a sequence of images $\left\{ {{\mathbf{I}_i}} \right\}_{i = 1}^N$ captured from a fixed trajectory with a narrow field of view, the task is to reconstruct the scene geometry and predict collision-free 6-DoF grasps. 

%The problem is inherently ill-posed due to sparse and limited views. However, NeuGrasp addresses this by leveraging background priors for end-to-end reconstruction and grasp prediction without explicit geometry supervision.

%===================Qingyu=============NeuGrasp takes as input scene images $\left\{ {{\mathbf{I}_i}} \right\}_{i = 1}^N$, background images $\left\{ {{\mathbf{B}_i}} \right\}_{i = 1}^N$, and camera parameters $\left\{ {{\mathbf{P}_i}} \right\}_{i = 1}^N$, reconstructing an SDF volume $\mathbf{S}$ of the scene to generate 6-DoF grasp predictions $\left\{ {{g_j}|{g_j} = \left( {{\mathbf{t}_j},{\mathbf{r}_j},{w_j},{q_j}} \right)} \right\}$, where $\mathbf{t}$ is the grasp center, $\mathbf{r}$ the orientation, $q$ the quality score, and $w$ the opening width, defined as in \cite{breyer2021volumetric}. End-to-end differentiable training enables generalization and fast inference for unseen scenes.

NeuGrasp takes as input the scene images $\left\{ {{\mathbf{I}_i}} \right\}_{i = 1}^N$, background images $\left\{ {{\mathbf{B}_i}} \right\}_{i = 1}^N$, and camera parameters $\left\{ {{\mathbf{P}_i}} \right\}_{i = 1}^N$, reconstructing an SDF volume $\mathbf{S}$ of the scene to generate 6-DoF grasp predictions $\left\{ {{g_j}|{g_j} = \left( {{\mathbf{t}_j},{\mathbf{r}_j},{q_j},{w_j}} \right)} \right\}$ defined by the grasp center $\mathbf{t} \in \mathbb{R}^3 $, the orientation $\mathbf{r} \in SO\left( 3 \right)$, the quality score $q \in \left[ {0,1} \right]$ and the opening width $w \in \mathbb{R}$. 

%where $\mathbf{t_j}$ is the grasp center, $\mathbf{r_j}$ is the orientation, $q_j$ is the quality score, and $w_j$ is the opening width.

%, as defined in \cite{breyer2021volumetric}. The end-to-end differentiable training enables generalization and fast inference for unseen scenes.

\subsection{Preliminaries}
% 可泛化的神经表面重建旨在通过神经渲染的方式利用一组输入元组恢复场景的几何（一堆公式）。现有的通用框架均先通过编码器获得特征图。遵循体积渲染的范式，沿相机原点到像素的光线由粗到细在近到远平面间采样并获得每一个采样点的位置。将光线上的采样点投影回特征图获得每视图的投影特征，使用网络聚合多视角特征预测采样点位置的sdf和颜色。之后利用neus所建模的数学函数将sdf转换为权重，沿光线与采样点的预测颜色进行体积以渲染获得采样光线的颜色，公式化如下：
% 可泛化的神经表面重建基于隐式神经表示的体积渲染学习范式，能够捕捉光线在不同材质表面上的传播和变化，能够有效感知非朗伯表面物体，从而允许重建透明和反光物体的几何。并且，因为体积渲染是自然可微的，这使得其具有与下游任务比如抓取相结合进行多任务学习的能力。

%================Qingyu ========Generalizable neural surface reconstruction recovers scene geometry using a neural rendering pipeline with multi-view inputs $\left\{ {{\mathbf{I}_j},{\mathbf{P}_j}} \right\}_{j=1}^N$, where $ {\mathbf{I}_j}$ are images and $ {\mathbf{P}_j}$ are camera parameters. The framework uses an encoder $\mathcal{E}$ to extract image features $\big\{{\mathbf{f}_{j}^{img}}\big\}_{j=1}^{N}$. Following volume rendering, $K$ points are sampled along a ray $\mathbf{r}$ from the camera center to the pixel. The geometry network $\mathcal{F}_{geo}$ and weight network $\mathcal{F}_{weight}$ estimate the SDF $\sigma_i$ and blending weights to compute radiance $\mathbf{c}_i$, by aggregating multi-view features and colors at each point $\mathbf{p}_i$ via projection and bilinear interpolation.

The generalizable neural surface reconstruction aims to recover the scene geometry using a neural rendering pipeline with multi-view inputs $\left\{ {{\mathbf{I}_j},{\mathbf{P}_j}} \right\}_{j=1}^N$, where $ {\mathbf{I}_j}$ and $ {\mathbf{P}_j}$ are images and camera parameters, respectively. The pipeline uses an encoder $\mathcal{E}$ to extract image features $\big\{{\mathbf{f}_{j}^{img}}\big\}_{j=1}^{N}$. In volume rendering, $K$ points are sampled along a ray $\mathbf{r}$ from the camera center to the pixel. At each point $\mathbf{p}_i$, the geometry network $\mathcal{F}_{geo}$ aggregates multi-view features via projection and bilinear interpolation to estimate the SDF $\sigma_i$, while the weight network $\mathcal{F}_{weight}$ aggregates multi-view features to estimate the blending weights, which are then multiplied by colors and summed to compute the radiance $\mathbf{c}_i$.

% In volume rendering, $K$ points are sampled along the ray $\mathbf{r}$ from the camera center to the pixel, where a weight function is learned for each point to obtain the SDF representation. The geometry network $\mathcal{F}_{geo}$ and weight network $\mathcal{F}_{weight}$ are utilized to aggregate multi-view features at each sampled point $\mathbf{p}_i$ via projection and bilinear interpolation.

%estimate the SDF $\sigma_i$ and blending weights to compute the radiance $\mathbf{c}_i$, by aggregating multi-view features and colors at each point $\mathbf{p}_i$ via projection and bilinear interpolation.

The predicted SDF is converted into weights along the ray using the conversion function from \cite{wang2021neus}, which are then applied for volume rendering.

%=============Qingyu============The predicted SDF converts to weights using the function modeled in \cite{wang2021neus}, which are then used for volume rendering along the ray with $K$ points, following the equation in \cite{wang2021neus} for accumulated transmittance and opacity values.

\subsection{Reconstruction with Background Priors} 

\subsubsection{Reconstruction Transformers}
%====================Qingyu =================As in \cite{ren2023volrecon}, we use Transformer architecture to fuse multi-view features and predict geometry and blending weights, capitalizing on its superior information capture effectiveness. Additionally, we integrate a prior volume inferred from multi-view features to provide global spatial information.

As in \cite{ren2023volrecon}, we use Transformers to integrate multi-view features and predict geometry and weights for blending. Additionally, we introduce a high-level shape prior volume obtained from multi-view features to provide global shape prior. 

\textbf{High-level Shape Prior Volume}. The mean and variance of multi-view features at each voxel center are calculated in the workspace volume. To handle the inconsistencies on transparent and specular surfaces, we apply a 3D U-Net to infer the shape details, producing a high-level prior volume $\mathbf{V}$ that represents the global prior.

%=====================Qingyu===============\textbf{High-level Prior Volume}. We calculate the mean and variance of multi-view features at each position in the workspace volume. To handle inconsistencies on transparent and specular surfaces, we apply a 3D U-Net to infer spatial information, producing a high-level feature volume $\mathbf{V}$ with global shape priors.

%================Qingyu \textbf{View Transformer with Feature Fusion}. A view-fusion transformer $\mathcal{T}_{view}$ combines multi-view features $\big\{ \mathbf{f}_{j}^{img} \big\}_{j=1}^{N}$ with the volume feature $\mathbf{f}_v$ at position $\mathbf{p}$, obtained via trilinear interpolation in $\mathbf{V}$. Following \cite{ren2023volrecon}, \cite{na2024uforecon}, a learnable query token $\mathbf{f}_t$ is introduced to encapsulate the unified view feature $\mathbf{f}_{u}$. All features are combined using linear self-attention transformers \cite{sun2021loftr}:
\textbf{View Transformer with Feature Fusion}. The view-fusion transformer $\mathcal{T}_{view}$ combines multi-view features $\big\{ \mathbf{f}_{j}^{img} \big\}_{j=1}^{N}$ with volume feature $\mathbf{f}_v$ at position $\mathbf{p}$, which is obtained through trilinear interpolation in $\mathbf{V}$. Following \cite{ren2023volrecon}, \cite{na2024uforecon}, a learnable query token $\mathbf{f}_t$ is introduced to obtain the unified view feature $\mathbf{f}_{u}$. These features are then combined using linear self-attention transformers \cite{sun2021loftr}:
$$
{\mathbf{f}_u},\big\{ \Tilde{\mathbf{f}}^{img}_{j} \big\}^{N}_{j=1} = \mathcal{T}_{view}\bigl( \big\{ \mathbf{f}^{img}_{j} \left( \pi \left( \mathbf{p} \right) \right)\big\}^{N}_{j=1},{\mathbf{f}_v},{\mathbf{f}_t} \bigr), \eqno{(5)}
$$
where $\Tilde{\mathbf{f}}^{img}$ denotes the attention-weighted feature used for color blending and $\pi \left( \cdot \right)$ is the projection function.

%============Qingyu=======where $\Tilde{\mathbf{f}}^{img}$ represents the reintegrated feature for rendering and $\pi \left( \cdot \right)$ denotes perspective projection.

%===========Qingyu=======\textbf{Ray Transformer with Geometric Awareness}. Given the non-local nature of SDF, we embed the workspace volume coordinates using fixed positional encoding, and compress them with the unified view feature via an MLP. The sampling order along the ray reveals occlusion information, so a ray transformer with linear self-attention is applied to consolidate spatial information and redistribute attention along the ray:

\textbf{Ray Transformer with Geometric Awareness}. Given the non-local characteristic of SDF, we embed the workspace volume coordinates using fixed positional encoding. These embeddings are then compressed with the unified view feature through an MLP. The sampling order along the ray indicates the occlusion information. Therefore, we apply a ray transformer with linear self-attention to integrate spatial information and adjust attention along the ray:
$$
{\mathbf{f}_{geo}} = \mathcal{T}_{ray}\left( \operatorname{MLP}\left( {\mathbf{f}_u},{\gamma _{coord}} \right),{\gamma _{order}} \right). \eqno{(6)}
$$
Here, $\gamma$ represents positional encoding, and $\mathbf{f}_{geo}$ is the geometry feature used to decode the SDF.

\subsubsection{Background Prior Based Feature Enhancement}

% 尽管上述结构已经可以取得一定成效，但是透明和反光物体复杂的光学特性使得模型（仍然遭受了错误的表面估计）很难捕捉正确的表面位置关系。由于在机器人抓取任务中场景的背景始终保持不变，我们充分利用了场景的背景信息，使得模型能够更好地关注于前景物体，从而提升对透明和反光物体表面的感知能力。
% 背景减除旨在从视频序列或图像中分离出前景目标。这种技术通过建立和更新背景模型，进而从当前帧中减去背景，从而提取出前景物体
%============Qingyu=============Although effective, the complex optical properties of transparent and specular objects make accurate surface estimation challenging. Since the surrounding environment and tabletop background remain static during robotic grasping, we leverage this background information to guide the model's focus on foreground objects, improving perception of transparent and specular objects.

Although the aforementioned reconstruction pipeline is effective in capturing the geometry of diffuse surfaces, accurately estimating the surfaces of transparent and specular objects remains challenging due to their complex optical properties. 

In many grasping applications, the background scenarios remain static during grasping. Inspired by the background subtraction method  \cite{piccardi2004background}, we leverage this background prior to guide the attention of the reconstruction module on foreground objects rather than background information. Our approach allows the model to notice subtle differences between the transparent and specular surfaces and the background.
%, thereby improving the perception of transparent and specular objects.

%target objects are better attended to, even for transparent materials.

To reduce the effect of noise to illumination variations, instead of performing direct background subtraction, we first employ a lightweight Res U-Net \cite{ronneberger2015u} to extract features from scene images $\big\{{\mathbf{f}_{j}^{scn}}\big\}_{j=1}^{N}$ and background images $\big\{{\mathbf{f}_{j}^{bg}}\big\}_{j=1}^{N}$. Next, we propose residual feature enhancement based on subtraction attention mechanism \cite{zhao2021point} to re-weight the original features to obtain foreground-attentive features $\big\{{\mathbf{f}_{j}^{attn}}\big\}_{j=1}^{N} \in \mathbb{R}^{\frac{H}{4} \times \frac{W}{4} \times C}$. 
$$
\mathbf{f}^{attn} = \rho\left( f\left( \eta\left( \mathbf{f}^{scn} \right) - \varphi \left( \mathbf{f}^{bg} \right) \right) \right) \odot \delta \left( \mathbf{f}^{scn} \right), \eqno{(7)}
$$where $\eta$, $\varphi$, and $\delta$ are linear projections, $\rho$ is the sigmoid normalization function, and $f$ is a mapping function. It is observed from experiments that with the residual feature enhancement, the reconstruction demonstrates improved attention to foreground objects, which produces  more accurate surface reconstruction results for grasping.

\subsubsection{Spatial Residual Information Aggregation}
The residual features, obtained by subtraction of background features from scene features, implicitly indicates the occupancy of foreground objects. 
Building on the aforementioned prior volume, we extend the residual features from 2D feature maps to 3D feature volumes to obtain a prior volume that encapsulates  global occupancy information. This occupancy prior imposes a more explicit and powerful constraint on spatial perception. Consequently, we combine the foreground-attentive features and prior knowledge from occupancy feature volume with a query token using another view transformer.

To avoid confusion, we restate the symbols as follows. The shape-prior volume and occupancy-prior volume are represented as $\mathbf{V}_{shp}$ and $\mathbf{V}_{occ}$, respectively. The volume features obtained through trilinear interpolation are denoted as $\mathbf{f}_{v}^{shp}$ and $\mathbf{f}_{v}^{occ}$. The view transformer that aggregates source information is denoted as $\mathcal{T}_{view}^{src}$, while the view transformer that aggregates residual information is denoted as $\mathcal{T}_{view}^{res}$:
$$
{\mathbf{f}_{u}^{src}},\big\{{\Tilde{\mathbf{f}}^{src}_{j}}\big\}^{N}_{j=1} = {\mathcal{T}_{view}^{src}}\bigl( {\big\{{{\mathbf{f}}^{scn}_{j}}{\scriptstyle {\left( {\pi \left( \mathbf{p} \right)} \right)}}\big\}^{N}_{j=1}},{\mathbf{f}_{v}^{shp},{\mathbf{f}_{t}^{src}}} \bigr), \eqno{(8)}
$$
$$
{\mathbf{f}_{u}^{res}},\big\{{\Tilde{\mathbf{f}}^{res}_{j}}\big\}^{N}_{j=1} = {\mathcal{T}_{view}^{res}}\bigl( {\big\{{{\mathbf{f}}^{attn}_{j}}{\scriptstyle {\left( {\pi \left( \mathbf{p} \right)} \right)}\big\}^{N}_{j=1}}},{\mathbf{f}_{v}^{occ},{\mathbf{f}_{t}^{res}}} \bigr). \eqno{(9)}
$$
%=================Qingyu======Subsequently, we employ an MLP to fuse two types of unified view features, $\mathbf{f}_{u}^{src}$ and $\mathbf{f}_{u}^{res}$. Additionally, a 5-layer MLP is utilized to combine two types of reintegrated features and predict blending weights.

Subsequently, we employ an MLP to fuse the unified view features, $\mathbf{f}_{u}^{src}$ and $\mathbf{f}_{u}^{res}$. Additionally, a 5-layer MLP is used to fuse two types of attention-weighted features and predict the weights for blending.

\subsection{Grasp Detection with Reconstructed Geometry}
% 有了重建的场景隐式几何，我们先通过查询得到一个离散化的体素网格。由于网格的规则性，我们将tsdf voxel grid输入一个3D CNN网络，将每一个体素中心作为抓取查询位置生成相应的抓取表示参数，inspired by VGN。这种输入完整场景信息的方式避免了物体掩码或者显式碰撞检测的需要。同时，从几何体积到抓取体积的映射可以自然地与前述重建流程相拼接进行端到端的训练。
%============================Qingyu===========With the reconstructed implicit geometry, we first use volumetric queries to obtain a discrete volume grid. Leveraging its regularity, we input the TSDF voxel grid into a 3D CNN to generate grasp parameters for each voxel center, inspired by \cite{breyer2021volumetric}. This approach eliminates the need for a target mask \cite{sundermeyer2021contact} or explicit collision detection \cite{fang2020graspnet}. The geometric-to-grasp volume mapping integrates seamlessly with the reconstruction pipeline, enabling end-to-end, differentiable training for efficient optimization. Finally, we apply mask-out and non-maxima suppression, following \cite{breyer2021volumetric}, to ensure reliable grasp candidates.

With the reconstructed implicit geometry, we first use volumetric queries to obtain a discrete volume grid. Leveraging its regularity, we feed the TSDF voxel grid into a 3D CNN to generate grasp parameters for each voxel center \cite{breyer2021volumetric}. In our approach, the segmentation mask of foreground objects \cite{sundermeyer2021contact}  or explicit collision detection \cite{fang2020graspnet} is not necessary. The volumetric mapping from geometry to grasp parameters integrates seamlessly with the reconstruction pipeline, enabling end-to-end, differentiable training for efficient optimization. Finally, we apply mask-out and non-maxima suppression as in \cite{breyer2021volumetric} to ensure reliable grasp predictions.

\subsection{Implementation Details}

\subsubsection{Camera Viewpoint Trajectory}

%=======================Qingyu=======Adequate scene observation improves geometric reconstruction and grasp prediction but reduces versatility. GraspNeRF \cite{dai2023graspnerf} used six $360\degree$ views, while RGBGrasp \cite{liu2024rgbgrasp} employed a $90\degree$ cylindrical path with 12 dense views. To balance accuracy and versatility, we propose a challenging viewpoint trajectory. 

While dense scene observation improves geometric reconstruction and grasp prediction, it reduces flexibility in real-world scenarios. For example, GraspNeRF \cite{dai2023graspnerf} uses six $360\degree$ views, and RGBGrasp \cite{liu2024rgbgrasp} uses a $90\degree$ cylindrical path with 12 dense views. 

%To achieve a balance between accuracy and versatility, we propose a challenging viewpoint trajectory that aims to optimize both performance and flexibility.

We sample 4 camera views along a spiral trajectory that covers one-sixth of a hemisphere with spherical coordinates: radius $r \in \left[0.4, 0.5\right]$ m, polar angle $\theta \in \mathcal{U}\left(\frac{\pi}{12}, \frac{\pi}{8}\right)$, and azimuthal angle $\phi \in \mathcal{U}\left(0, \frac{\pi}{3}\right)$, aiming to optimize both performance and flexibility.

%with spherical coordinates: radius $r \in \left[0.4, 0.5\right]$ m, polar angle $\theta \in \mathcal{U}\left(\frac{\pi}{12}, \frac{\pi}{8}\right)$, and azimuthal angle $\phi \in \mathcal{U}\left(0, \frac{\pi}{3}\right)$. This sparse yet flexible trajectory enhances 

%================Qingyu========We sample 4 camera views along a spiral trajectory covering one-sixth of a hemisphere, with spherical coordinates: radius $r \in \left[0.4, 0.5\right]$ m, polar angle $\theta \in \mathcal{U}\left(\frac{\pi}{12}, \frac{\pi}{8}\right)$, and azimuthal angle $\phi \in \mathcal{U}\left(0, \frac{\pi}{3}\right)$. This sparse yet flexible trajectory enhances efficiency and versatility.

\subsubsection{Training Details}

%We use end-to-end training to jointly learn reconstruction and grasping. The total loss is a weighted sum of these losses.The loss functions are:

We use end-to-end training to jointly optimize both reconstruction and grasping tasks. The total loss is a weighted sum of the following losses:

\textbf{Photometric Loss.} We compute the mean squared error (MSE) between rendered and ground truth pixel colors over sampled rays \cite{mildenhall2020nerf}.

%======Qingyu===========\textbf{Photometric Loss.} We compute MSE between rendered and ground truth pixel colors over sampled rays, as in \cite{mildenhall2020nerf}.

%===========Qingyu======\textbf{Grasping Loss.} Based on \cite{breyer2021volumetric}, we supervise grasp quality, orientation, and gripper width using binary cross-entropy, L2, and quaternion losses, with adjustments for gripper symmetry. Supervision is applied only for successful grasps.

\textbf{Grasping Loss.} Following \cite{breyer2021volumetric}, we supervise grasp quality, orientation, and gripper width. Supervision is applied only to successful grasps.

\textbf{Regularization Loss.} We include an Eikonal term to enforce correct SDF properties, ensuring gradient magnitudes are close to 1 as in \cite{wang2021neus}.

We sample 2304 rays per batch using a learning rate of 1e-4 with exponential decay. All experiments are run on an NVIDIA RTX 4090D GPU.

\section{EXPERIMENTS}

%=====================Qingyu==============In this section, we evaluate our approach through experiments in both simulated and real-world robotic environments, along with ablation studies using simulation data to assess key components.

In this section, we evaluate our approach in both simulated and real-world experiments. Ablation studies are also performed to evaluate the impact of different components.

\begin{figure}[t]
    \centering
    \includegraphics[width=0.48\textwidth]{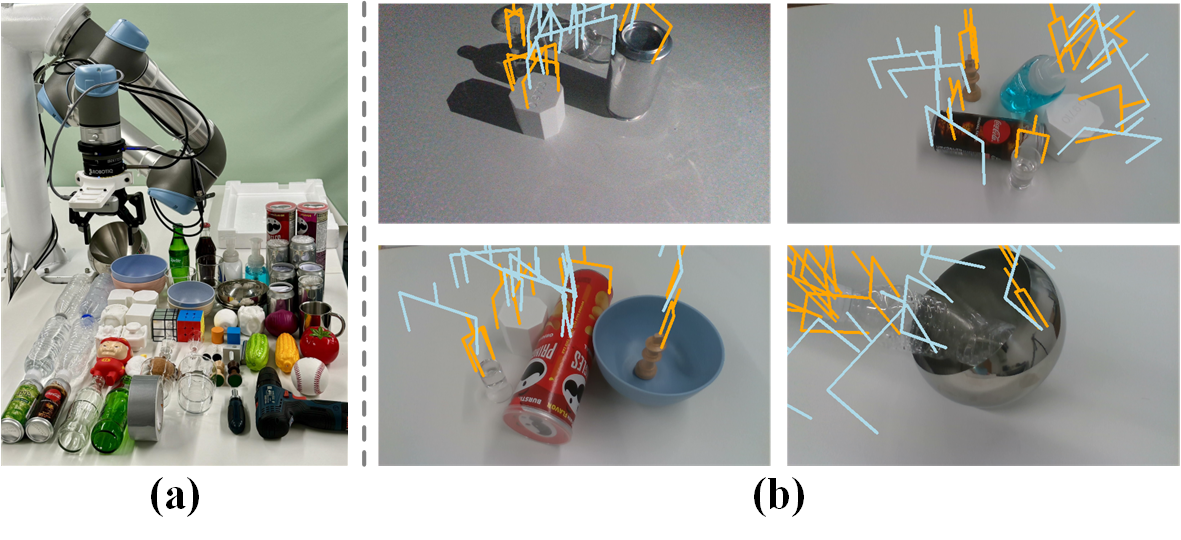}
    \vspace{-0.84cm}
    \caption{\textbf{Real-world Experiments.} (a) Experimental settings. (b) Visualization of grasps, where orange represents positive labels and light blue represents negative labels.}
    \label{fig3}
    \vspace{-0.5cm}
\end{figure}

\subsection{Experiment Setup}

\subsubsection{Simulation Setup}

%==================Qingyu============We use PyBullet \cite{coumans2016pybullet} for physics-based grasping simulation and Blender \cite{blender} to generate photorealistic synthetic data. Grasp data is autonomously collected by randomly sampling grasp centers and orientations near object surfaces, followed by executing these attempts in simulation. The workspace is a $30 \times 30 \times 30 \ \text{cm}^3$ tabletop area, where all objects are confined. We simulate three types of scenes:

We use PyBullet \cite{coumans2016pybullet} for physics-based grasping simulation and Blender \cite{blender} to generate photorealistic synthetic data. Grasp data is autonomously collected by randomly sampling grasp centers and orientations near object surfaces, followed by executing these attempts in the simulation. The workspace is a $30 \times 30 \times 30 \ \text{cm}^3$ tabletop area following \cite{breyer2021volumetric, dai2023graspnerf}. We simulate three types of scenes: a \textbf{Single} setting where individual objects are placed at the tabletop center, a \textbf{Pile} setting where multiple objects are dropped from above to form a cluttered pile, and a \textbf{Packed} setting multiple objects are placed upright in random positions to form a densely packed arrangement.

\begin{itemize}
    \item \textbf{Single}: Individual objects are placed in the workspace, each in an isolated and upright position.
    \item \textbf{Pile}: Multiple objects are dropped from above to form a cluttered pile.
    \item \textbf{Packed}: Multiple objects are placed upright in random positions, simulating a densely packed arrangement.
\end{itemize}

%==================Qingyu=========For baselines that requires depth information, the simulation environment also incorporates realistic depth sensor noise using a dedicated depth sensor simulator \cite{dai2022domain}, ensuring that simulated depths closely mirror real-world sensor characteristics.

For baselines that require depth information, the simulation environment also considers realistic depth noise using depth sensor simulator \cite{dai2022domain}. This ensures that the simulated depths accurately reflects real-world sensor characteristics.

\begin{table*}[t]
    \centering

    \caption{RESULTS OF SIMULATION EXPERIMENTS IN \textbf{PACKED} SCENE, \textbf{ MATERIAL}: TRANSPARENT\&SPECULAR 
 (DIFFUSE)}
    \label{sim_packed}
    \small
    
    \resizebox{0.8\textwidth}{!}{  % 调整表格宽度适应页面
    \begin{tabular}{l|cccccccc}
        \hline
        \multirow{2}{*}{Method}
        & \multicolumn{4}{c}{Selective Random} & \multicolumn{4}{c}{Top Score} \\
        \cline{2-5} \cline{6-9}
        & SR (\%) & FSR (\%) & DR (\%) & L1 & SR (\%) & FSR (\%) & DR (\%) & L1 \\
        \hline
        VGN & 27.4(47.9) & 21.0(40.5) & 18.0(36.7) & 0.273(0.205) & 12.7(44.5) & 10.0(46.2) & 7.3(27.8) & 0.288(0.222) \\
        SwinDR-VGN  & 40.8(66.0) & 43.0(57.8) & 35.6(55.8) & 0.239(0.204) & 46.0(46.0) & 40.0(39.5) & 39.1(38.1) & 0.240(0.234) \\
        \hline
        GraspNeRF w/ TSDF & \underline{84.2}(\underline{74.2}) & \underline{85.5}(\underline{79.9}) & \underline{80.3}(\underline{75.7}) & \textbf{0.110}(\textbf{0.112}) & \underline{76.3}(\underline{81.5}) & \underline{80.5}(\underline{93.5}) & \underline{76.6}(\underline{81.1}) & \textbf{0.113}(\textbf{0.112}) \\
        \hline
        GraspNeRF w/o TSDF & 47.6(43.3) & 54.0(52.8) & 50.2(43.9) & 0.894(0.897) & 50.4(46.9) & 64.5(61.4) & 44.8(43.1) & 0.895(0.897) \\
        RGBGrasp-GT & 36.2(42.7) & 23.0(40.0) & 18.9(33.1) & 0.807(0.745) & 25.5(36.3) & 14.5(42.0) & 12.9(22.4) & 0.811(0.733) \\
        \textbf{Ours (NeuGrasp)} & \textbf{86.3}(\textbf{88.3}) & \textbf{86.0}(\textbf{86.0}) & \textbf{81.0}(\textbf{81.0}) & \underline{0.153}(\underline{0.152}) & \textbf{86.8}(\textbf{90.3}) & \textbf{92.0}(\textbf{94.5}) & \textbf{79.1}(\textbf{82.9}) & \underline{0.155}(\underline{0.154}) \\
        \hline
    \end{tabular}
    }
    % \vspace{-0.5cm}
\end{table*}

\begin{table*}[t]
    \centering
        \vspace{-0.2cm}
    \caption{RESULTS OF REAL-WORLD EXPERIMENTS IN \textbf{PILE} AND \textbf{PACKED} SCENES,
             \textbf{MATERIAL}: MIXED}
    \label{real}
    \small
    \resizebox{0.8\textwidth}{!}{  % 调整表格宽度适应页面
    \begin{tabular}{l|cccccc}
        \hline
        \multirow{2}{*}{Method} & \multicolumn{3}{c}{Pile} & \multicolumn{3}{c}{Packed} \\
        \cline{2-7}
        & SR (\%) & FSR (\%) & DR (\%) & SR (\%) & FSR (\%) & DR (\%) \\
        \hline
        GraspNeRF w/ TSDF & 38.9 (28 / 72) & 40.0 (10 / 25) & 22.4 (28 / 125) & 46.7 (42 / 90) & 64.0 (16 / 25) & 33.6 (42 / 125) \\
        \textbf{Ours (NeuGrasp)} & \underline{63.7} (72 / 113) & \underline{68.0} (17 / 25) & \underline{57.6} (72 / 125) & \underline{84.1} (106 / 126) & \textbf{92.0} (23 / 25) & \underline{84.8} (106 / 125) \\
        \textbf{Ours (NeuGrasp-RA)} & \textbf{84.5} (120 / 142) & \textbf{92.0} (23 / 25) & \textbf{96.0} (120 / 125) & \textbf{85.8} (121 / 141) & \underline{88.0} (22 / 25) & \textbf{96.8} (121 / 125)  \\
        \hline
    \end{tabular}
    }
    % \vspace{-0.5cm}
\end{table*}

\begin{table*}[t]
    \centering
    \vspace{-0.2cm}
    \caption{ABLATION STUDIES IN \textbf{PILE} AND \textbf{PACKED} SCENES,
            \textbf{MATERIAL}: TRANSPARENT\&SPECULAR (DIFFUSE)}
    \label{ablation}
    \small
    \resizebox{0.8\textwidth}{!}{  % 调整表格宽度适应页面
    \begin{tabular}{l|cccccccc}
        \hline
        \multirow{2}{*}{Method} & \multicolumn{4}{c}{Pile} & \multicolumn{4}{c}{Packed} \\
        \cline{2-9}
        & SR (\%) & FSR (\%) & DR (\%) & L1 & SR (\%) & FSR (\%) & DR (\%) & L1 \\
        \hline
        w/o $\mathbf{f}^{attn}$ & 51.4(45.1) & 41.6(34.0) & 21.6(18.9) & 0.906(0.887) & 76.3(77.4) & 73.0(75.0) & 69.1(71.5) & 0.887(0.899) \\
        w/o $\mathbf{V}_{occ}$ & 61.8(65.1) & 65.0(63.0) & 42.6(42.5) & 0.360(0.379) & 82.2(78.5) & 83.0(67.5) & 78.1(73.6) & 0.321(0.326) \\
        w/o backgrounds & 58.3(58.8) & 62.5(62.5) & 44.4(41.4) & 0.403(0.414) & 79.6(76.8) & 82.0(85.5) & 78.5(77.4) & 0.378(0.375) \\
        \textbf{Ours (NeuGrasp)} & \textbf{65.2}(\textbf{65.3}) & \textbf{67.5}(\textbf{64.5}) & \textbf{45.6(43.2)} & \textbf{0.164}(\textbf{0.171}) & \textbf{86.3}(\textbf{88.3}) & \textbf{86.0}(\textbf{86.0}) & \textbf{81.0}(\textbf{81.0}) & \textbf{0.153}(\textbf{0.152}) \\
        \hline
    \end{tabular}
    }
    \vspace{-0.5cm}
\end{table*}

\subsubsection{Real-world Setup}

%==================Qingyu==========As shown in Fig. \ref{fig3}, we use a UR5 robot arm with a Robotiq 2-Finger 85 gripper for real-world grasping tasks. An Intel RealSense D435 RGB-D camera mounted on the robot's wrist captures RGB data, though the depth information is not used. The robot operates within the same $30 \times 30 \times 30 \ \text{cm}^3$ workspace, inferring grasp poses in real time after capturing 4 scene images along a predefined trajectory. Backgrounds are prepared in advance and reused across all scenes. To showcase the method’s depth-free supervision, we collected a real-world dataset of backgrounds, scenes, and grasp data, which was used to fine-tune the model, yielding the improved NeuGrasp-RA, significantly reducing the domain gap between simulation and reality.

As shown in Fig. \ref{fig3}, we use a UR5 robot arm with a Robotiq 2-Finger 85 gripper for real-world grasping tasks. An Intel RealSense D435 RGB-D camera, mounted on the robot's wrist, captures RGB data. The depth information is not used. The robot operates within a $30 \times 30 \times 30 \ \text{cm}^3$ workspace, inferring grasp poses in real time after capturing 4 scene images along the predefined trajectory. 

Backgrounds are pre-prepared and reused across all scenes. We also collect a real-world dataset consisting of backgrounds, scenes, and grasp data. This real-world dataset is used to fine-tune the model, resulting in an improved version of NeuGrasp, referred to as NeuGrasp-RA. NeuGrasp-RA generates more accurate grasp predictions by reducing the domain gap between simulation and real-world environments.

\subsubsection{Object Set}

To ensure consistency with baseline methods, we use 473 hand-scaled object meshes in  simulation, with 417 for training and 56 for testing \cite{breyer2021volumetric}. During training, we randomize object textures and materials, including transparent, specular, and diffuse types. Additionally, we collect 62 diverse household objects for real-world fine-tuning and evaluation. See Fig. \ref{fig3} for an overview of these objects.

%===============Qingyu====To ensure consistency with baseline methods, we use 473 hand-scaled object meshes in our simulation, with 417 for training and 56 for testing. During training, we randomize object textures and materials, including transparent, specular, and diffuse types. Additionally, we collect 62 diverse household objects for real-world fine-tuning and evaluation. See Fig. \ref{fig3} for an overview of these objects.

\subsection{Baseline Methods}
All the methods are evaluated using 4 camera views along the same trajectory, unless noted otherwise.

\subsubsection{RGB-D-Based Methods}

\begin{itemize}
    \item \textbf{VGN \cite{breyer2021volumetric}}: A volumetric grasp detection network that uses TSDF volumes from multiple depth views to predict 6-DoF grasp poses.
    \item \textbf{SwinDR-VGN}: A variant of VGN that incorporates SwinDRNet \cite{dai2022domain} for depth restoration to address depth inaccuracies with transparent and specular materials.
\end{itemize}

\subsubsection{RGB-TSDF-Based Method}

\begin{itemize}
    \item \textbf{GraspNeRF \cite{dai2023graspnerf}}: A real-time, multi-view 6-DoF grasp detection network utilizing generalizable NeRF. GraspNeRF is a material-agnostic grasping method.
\end{itemize}

\subsubsection{RGB-Based Method}

\begin{itemize}
    \item \textbf{RGBGrasp-GT}: RGBGrasp \cite{liu2024rgbgrasp} is a reconstruction-grasp pipeline leveraging Nerfacto \cite{tancik2023nerfstudio}  with pre-trained monocular depth estimator %prediction
    \cite{zhou2022devnet} and grasp detector %detection
    \cite{fang2023anygrasp}. In RGBGrasp-GT, we use ground truth depth maps from simulation instead of depth prediction from \cite{zhou2022devnet} to provide supervision and replace \cite{fang2023anygrasp} with VGN for grasp detection for a fair comparison.
\end{itemize}

\subsection{Evaluation Metrics}

%We use the following metrics for evaluation:

%================Qingyu======We measure our performance with the following evaluation metrics:
\begin{itemize}
    \item \textbf{Success Rate (SR)}: The ratio of successful grasps to the total number of grasp attempts.
    \item \textbf{Declutter Rate (DR)}: The ratio of objects successfully removed from the scene.
    \item \textbf{First Success Rate (FSR)}: The ratio of rounds with a successful first grasp to the total number of rounds.
    \item \textbf{Mean Absolute Error on TSDF (L1)}: The average absolute difference between the reconstructed surface and the ground truth.
\end{itemize}

\subsection{Simulated Grasping Experiments}

%=========================Qingyu===============In each scene, we conduct multiple decluttering experiments using two material sets: one with transparent and specular materials, and the other with diffuse materials. We perform 200 rounds in pile and packed scenarios, and 56 rounds in single-object scenarios, to demonstrate the material-agnostic capability of our method. Each pile and packed scene contains 5 objects. We use two grasp execution strategies: \textbf{1) Top Score}, executing the highest-scored grasp, and \textbf{2) Selective Random}, executing a random grasp with a score above 0.9. These strategies highlight both optimal performance and robustness by embracing diverse, high-quality grasp options. Each trial continues until the workspace is cleared, the system fails, or two consecutive grasp failures occur.

In each scene, we conduct multiple decluttering experiments with: 1) transparent and specular materials, and 2) diffuse materials. We perform 200 rounds for the pile and packed scenarios, and 56 rounds for single-object scenarios, to demonstrate the material-agnostic grasping capability of our method. Each pile and packed scene contains 5 objects. We use two grasp execution strategies: \textbf{1) Top Score}, which executes the highest-scored grasp, and \textbf{2) Selective Random}, which executes a random grasp with a score above 0.9, which considers more diverse grasp options. The robot continues executing grasps until the workspace is cleared, the prediction fails, or two consecutive grasp failures occur.

To ensure a fair comparison, we re-train GraspNeRF on our dataset using the viewpoint trajectory and parameter settings in \cite{dai2023graspnerf}. We train two GraspNeRF models: one with TSDF loss (\textbf{GraspNeRF w/ TSDF}) and one without (\textbf{GraspNeRF w/o TSDF}). Since VGN was trained with extensive data augmentation and on a significantly larger dataset, we directly use the best pre-trained VGN model \cite{breyer2021volumetric}. The results for the packed scene are summarized in Tab. \ref{sim_packed}. More  results are provided in the supplementary materials.
%======================Qingyu===============To ensure fair comparison, we re-train GraspNeRF on our dataset using the specified viewpoint trajectory and their hyperparameter settings. To highlight the challenge of lacking geometry supervision, we train two GraspNeRF models: one with TSDF loss (\textbf{GraspNeRF w/ TSDF}) and one without it (\textbf{GraspNeRF w/o TSDF}). Since VGN is trained with extensive data augmentation and a much larger dataset, we use the best pre-trained VGN model directly. The packed scene results are summarized in Tab. \ref{sim_packed}, with additional experimental results available in the supplementary materials.

%================Qingyu======Our method surpasses VGN and SwinDR-VGN in all metrics. VGN struggles with TSDF consistency due to sensor noise and missing data, especially on transparent or specular surfaces. SwinDR-VGN, though using depth restoration, over-smooths surfaces, distorting object sizes and lowering grasp accuracy. RGB-D based methods also perform poorly with diffuse materials in limited view conditions, highlighting the challenge of our trajectory.

It is observed in Tab. \ref{sim_packed} that VGN struggles to generate accurate grasp predictions due to overly noisy and missing data from the sensor. While SwinDR-VGN  uses depth restoration, it tends to over-smooth object surfaces. RGB-D based methods still have difficulty in dealing with diffuse materials under limited view conditions, which highlights the challenge of grasping from a narrow field of view. GraspNeRF w/ TSDF performs better compared with GraspNeRF w/o TSDF. The direct TSDF supervision helps the reconstruction to align closely with the ground truth. 

%==========================Qingyu=======GraspNeRF w/ TSDF performs well, particularly in surface reconstruction, due to direct TSDF supervision, which guides the model to align closely with the geometric representation. In contrast, GraspNeRF w/o TSDF performs poorly in the absence of direct geometric loss. However, NeuGrasp surpasses it in most grasp metrics, despite lacking explicit geometry supervision. This is due to NeuGrasp’s ability to effectively extract and integrate scene and background information, generating a rich representation. By relying solely on the mathematical relationship between SDF and volumetric rendering, NeuGrasp uncovers geometric structures from RGB data alone. Moreover, NeuGrasp's design focuses more effectively on foreground targets, enhancing geometric structure recovery and the relationship between geometry and grasp. To illustrate, we used t-SNE \cite{van2008visualizing} to visualize scene and foreground-attentive features, comparing them with GraspNeRF's results in Fig. \ref{fig4}. The results show that NeuGrasp, after residual feature enhancement, focuses more on grasp targets, better identifies object contours, and clearly distinguishes objects from the background.

\begin{figure}[t]
    \centering
    \includegraphics[width=0.48\textwidth]{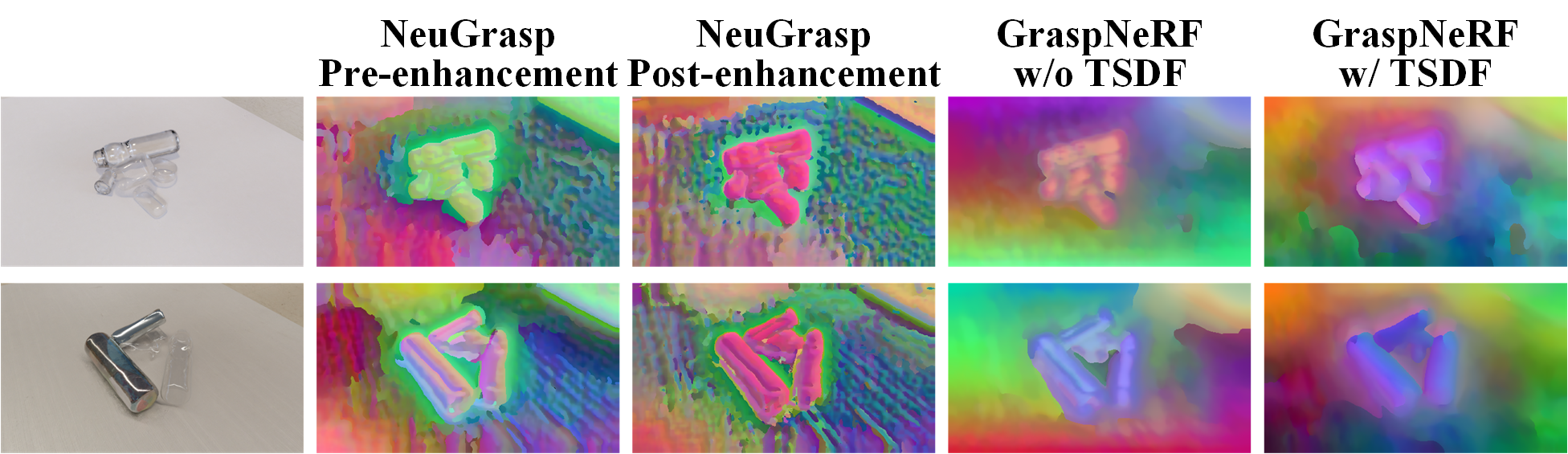}
    \caption{\textbf{Comparison of View Features between GraspNeRF and NeuGrasp.} It shows that with the residual feature enhancement, NeuGrasp improves focus on the foreground objects. The transparent and specular objects are clearly distinguished from the background.}
    \label{fig4}
    \vspace{-0.3cm}
\end{figure}

NeuGrasp outperforms all methods in most grasp metrics with no explicit geometry supervision. This is due to NeuGrasp’s ability to uncover geometric structures by effectively integrating scene and background information using only RGB data. Moreover, the residual feature enhancement in NeuGrasp not only facilitates the reconstruction process by better attending to foreground objects, resulting in more accurate surface reconstruction for grasping, but also sharpens the focus on target grasp region, leading to more precise and reliable grasp predictions. To illustrate, t-SNE \cite{van2008visualizing} is used to visualize scene and foreground-attentive features. The features of GraspNeRF are also shown in Fig. \ref{fig4}. The results show that with the residual feature enhancement, object contours can be clearly distinguished from the background, which validates the model's improved focus on target objects.

Due to narrow and sparse viewpoints, RGBGrasp-GT fails to predict accurate depth, even with ground truth depth supervision. Similar to classical NeRF methods, RGBGrasp-GT relies heavily on multiple views. The sparse views result in large depth variances, which results in inaccurate TSDF reconstruction results and unreliable grasp predictions. Moreover, RGBGrasp-GT requires retraining for each grasp, with an average prediction time of \textbf{56.7} seconds. In contrast, our method achieves an average inference time of \textbf{0.274} seconds, which demonstrates its advantages in real applications.

%    \caption{\textbf{Comparison of Visualizations across the Same Scene.} This comparison demonstrates that NeuGrasp achieves a cleaner reconstruction and offers greater grasp diversity compared to GraspNeRF. NeuGrasp-RA, fine-tuned with a small-scale real-world dataset, exhibits further enhanced performance, particularly in grasp diversity. }

\begin{figure}[t]
    \centering
    \includegraphics[width=0.48\textwidth]{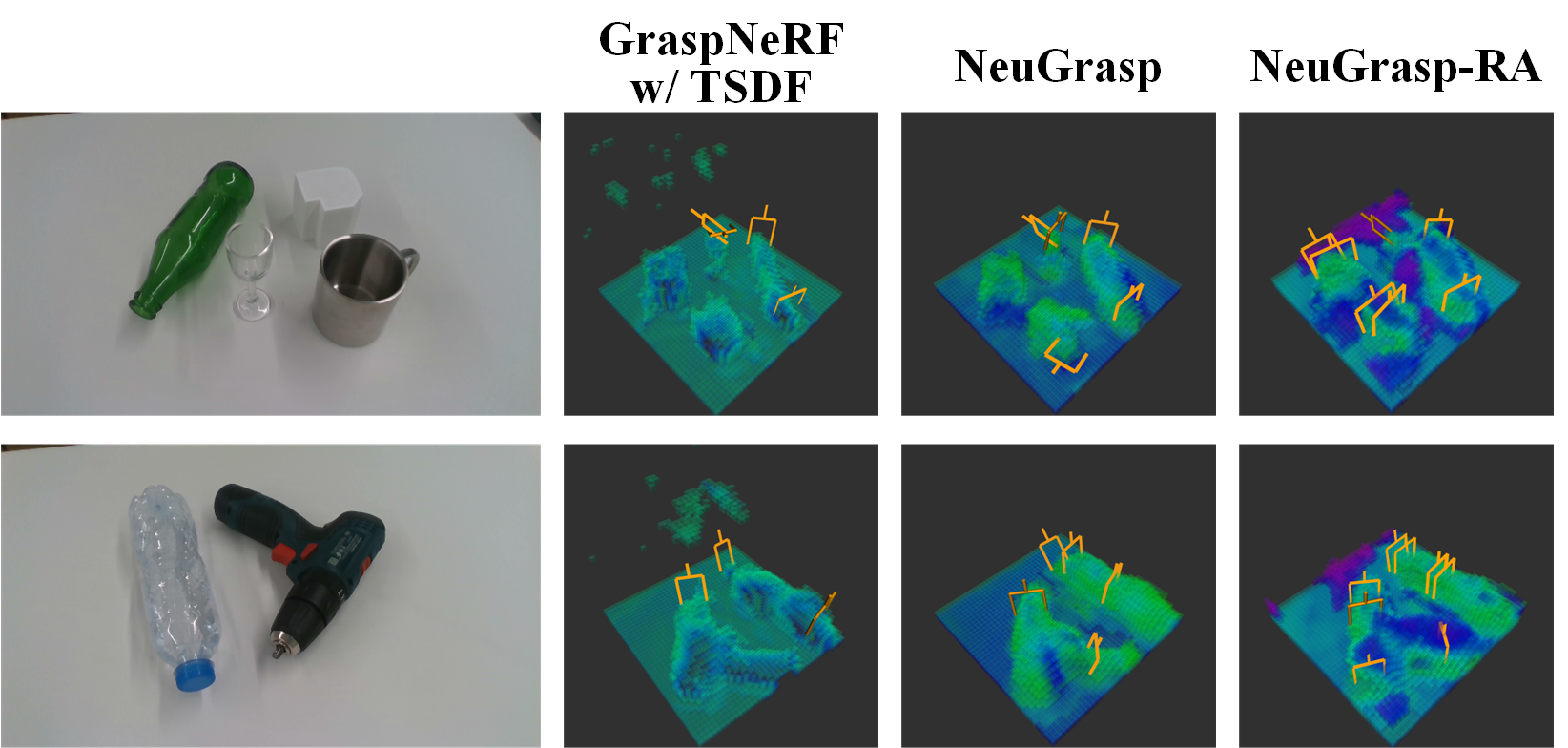}
    \caption{\textbf{Visualization Results.} NeuGrasp achieves a cleaner reconstruction and greater grasp diversity compared to GraspNeRF. NeuGrasp-RA, fine-tuned with a small-scale real-world data, further improves the performance, particularly in grasp diversity. }
    \label{fig5}
    \vspace{-0.5cm}
\end{figure}

\subsection{Real-world Grasping Experiments}
% 为了进一步探究真实世界场景下的抓取性能，我们收集了62个daily objects,进行了30轮的抓取实验在pile和packed的场景下。实验遵循了仿真实验的章程，并且为了验证鲁棒性，只采用Selective Random策略进行抓取执行。由于RGB-D based method的糟糕表现，以及真实世界下无法直接获取深度真值，所以我们只与GraspNeRF进行对比。为了突出我们轨迹的高效性，我们在真机实验中也使用了原文视角配置下的GraspNeRF，并使用其提供的最优模型。并且，为了突出论证不需要几何监督的好处，我们还利用收集到的物体中的一小部分构建了一个小规模的数据集，用其对模型进行微调。注意我们测试时使用的物体并不包括微调所使用的物体。

%==================Qingyu====To evaluate real-world grasping performance, we collected 62 daily objects and conducted 25 rounds of experiments in both pile and packed scenes. Following the same protocol as in simulations, we used only the Selective Random strategy for robustness. Given the poor performance of RGB-D methods and the lack of ground truth depth, we compared our method solely with GraspNeRF w/ TSDF. Additionally, we fine-tuned our model on a real dataset collected from a subset of objects, excluding the test objects.

To evaluate the performance of real-world grasping, we collect 62 daily objects and conduct 25 rounds of experiments in both pile and packed scenes. Using the same experimental setting as in simulations, we apply only the Selective Random strategy here. Due to the absence of ground truth depth maps, we compare our method  with GraspNeRF w/ TSDF. Additionally, we fine-tune our model on a small real-world dataset collected from a subset of objects,  where test objects are not included. 
%===========Qingyu==============Tab. \ref{real} shows NeuGrasp-RA achieving the best performance by eliminating the domain gap. Both of our methods significantly outperformed GraspNeRF w/ TSDF, showcasing their robustness and efficiency, despite GraspNeRF using geometry ground truth and domain randomization. Fig. \ref{fig5} visually demonstrates that our method produces cleaner scene geometry and more accurate grasp candidates. Notably, even with fine-tuning on a very small dataset, our approach effectively closes the domain gap and learns an inductive bias related to reachability, which significantly reduces planning failures and produces more reachable grasps, greatly improving grasping efficiency.

Tab. \ref{real} shows NeuGrasp-RA achieves the best performance. Fine-tuning on a small real-world dataset allows NeuGrasp to better adapt to real-world conditions, effectively reducing the domain gap. Although GraspNeRF uses geometry ground truth and domain randomization, NeuGrasp and NeuGrasp-RA significantly outperform GraspNeRF w/ TSDF. It can be observed from Fig. \ref{fig5} that our method produces cleaner scene reconstruction and more accurate grasp candidates. 

\subsection{Ablation Studies}
% 为了评估NeuGrasp每一部分的影响，我们在pile和packed场景下进行了消融研究 in simulation. 由于其综合性，我们使用了Selective Random策略。 
% Residual Feature Enhancement. 我们移除了残差特征提升模块，使用f代替fattn. 通过结果可以看出，性能在抓取方面大幅衰减，并且在重建方面直接崩坏。缺少该模块会导致模型无法合理利用背景信息（有效捕捉区分度较低的物体表面特征，例如玻璃表面）。同时过度的特征表示Over-representation of source features使得模型过度拟合，集中于一些错误且无关的细节，例如玻璃或反光表面的光照不一致。
% Occupancy Volume. 我们比对不使用occupancy-prior volume及其相关特征的模型，发现在pile场景下抓取性能相似，但在packed场景抓取性能略有下降，并且两者重建表现都下降很大。尽管由源特征构建的shape-prior volume在某种程度上提供了空间约束，但由于其复杂性，效果不如occupancy-prior volume那样直接有力。因此，在遮挡较为严重的packed场景下，抓取性能会受到影响。此外，由于缺乏显式的几何监督，简洁且明确的occupancy-prior volume显得尤为重要，因为它能够使模型直接关注于重建区域。
% Backgound priors.最后我们移除了所有与background相关的模块与特征通道。合适的模型大小与特征维度使其摒弃了过拟合的影响，在抓取性能方面有所回升，但还是低于合理利用背景信息的NeuGras w/o和full model，并且在重建性能方面相差很多。这证明了背景先验信息在杂乱物体抓取场景下的重要作用（有效性）。
%================Qingyu===============To evaluate the impact of each NeuGrasp component, we conducted ablation studies in simulated pile and packed scenes, as shown in Tab. \ref{ablation}, using the Selective Random strategy for comprehensiveness.

To evaluate the impact of each component in NeuGrasp, we conduct ablation studies in simulated pile and packed scenes (Tab. \ref{ablation}) using the Selective Random strategy.

\textbf{Residual Feature Enhancement.} It can be observed in Tab. \ref{ablation} that replacing $\mathbf{f}^{attn}$ with $\mathbf{f}^{scn}$ leads to a significant drop in grasp performance and a large L1 error. Without this module, the model fails to leverage background information, resulting in a collapse in reconstruction.

%and overfit irrelevant details like inconsistent lighting on glass or reflective surfaces.

%==================Qingyu============\textbf{Residual Feature Enhancement.} Removing this module, replacing $\mathbf{f}^{attn}$ with $\mathbf{f}^{scn}$, led to a significant decline in grasp performance and a complete collapse in reconstruction quality. Without this module, the model failed to leverage background information and overfit irrelevant details like inconsistent lighting on glass or reflective surfaces

\textbf{Occupancy-Prior Volume.} Removing this volume slightly reduces grasp performance in pile scenes but causes a significant drop in packed scenes and reconstruction quality overall. The occupancy-prior volume proves crucial for handling occluded packed scenes without explicit geometric supervision.

%================Qingyu=========\textbf{Occupancy-Prior Volume.} Omitting this volume slightly reduced grasp performance in pile scenes but caused a significant drop in grasp performance in packed scenes and reconstruction quality in both scene types. While the shape-prior volume offers some spatial constraints, it's less effective than the simpler and more powerful occupancy-prior volume, which is crucial in occluded packed scenes and in the absence of explicit geometric supervision.

%======================Qingyu--------------\textbf{Background Priors.} We removed all modules and features related to background priors. While adjusting model size and feature dimensions reduced overfitting and slightly improved grasping performance, it still fell short of NeuGrasp w/o $\mathbf{V}_{occ}$ and the full model, both of which leverage background information. Reconstruction performance also dropped significantly, highlighting the crucial role of background priors in cluttered object grasping.

\textbf{Background Priors.} When all modules related to background priors are removed, the model underperforms compared to both NeuGrasp w/o $\mathbf{V}_{occ}$ and the full NeuGrasp, both of which leverage background priors. The reconstruction performance also drops significantly, highlighting the crucial role of background priors in effective object grasping in cluttered scenes.

\section{CONCLUSIONS}
%=================Qingyu=============In this paper, we introduced NeuGrasp, a neural surface reconstruction method that enhances robotic grasping in cluttered environments, particularly with transparent and specular objects. By leveraging background priors and using transformers for multi-view feature aggregation, NeuGrasp enables robust, material-agnostic grasp detection without depth-related supervision, even in challenging conditions. Experimental results show its superior performance over state-of-the-art methods in both simulated and real-world settings. Future work will explore more generalized approaches to utilizing background priors, advancing general AI for versatile robotic manipulation.

In this paper, we introduced NeuGrasp, a neural surface reconstruction method for robotic grasping in cluttered scenes  with transparent and specular objects. By leveraging background priors and multi-view feature aggregation, NeuGrasp enables robust, material-agnostic grasp detection without depth-related supervision. Experimental results show its superior performance over state-of-the-art methods in both simulated and real-world scenarios. %Future work will explore more generalized approaches to utilizing background priors.


\bibliographystyle{IEEEtran}
\bibliography{IEEEabrv, IEEEexample, mybibfile} %, mybibfile}

\begin{thebibliography}{10}
\providecommand{\url}[1]{#1}
\csname url@samestyle\endcsname
\providecommand{\newblock}{\relax}
\providecommand{\bibinfo}[2]{#2}
\providecommand{\BIBentrySTDinterwordspacing}{\spaceskip=0pt\relax}
\providecommand{\BIBentryALTinterwordstretchfactor}{4}
\providecommand{\BIBentryALTinterwordspacing}{\spaceskip=\fontdimen2\font plus
\BIBentryALTinterwordstretchfactor\fontdimen3\font minus \fontdimen4\font\relax}
\providecommand{\BIBforeignlanguage}[2]{{%
\expandafter\ifx\csname l@#1\endcsname\relax
\typeout{** WARNING: IEEEtran.bst: No hyphenation pattern has been}%
\typeout{** loaded for the language `#1'. Using the pattern for}%
\typeout{** the default language instead.}%
\else
\language=\csname l@#1\endcsname
\fi
#2}}
\providecommand{\BIBdecl}{\relax}
\BIBdecl

\bibitem{fang2020graspnet}
H.-S. Fang, C.~Wang, M.~Gou, and C.~Lu, ``Graspnet-1billion: A large-scale benchmark for general object grasping,'' in \emph{Proceedings of the IEEE/CVF conference on computer vision and pattern recognition}, 2020, pp. 11\,444--11\,453.

\bibitem{sundermeyer2021contact}
M.~Sundermeyer, A.~Mousavian, R.~Triebel, and D.~Fox, ``Contact-graspnet: Efficient 6-dof grasp generation in cluttered scenes,'' in \emph{2021 IEEE International Conference on Robotics and Automation (ICRA)}.\hskip 1em plus 0.5em minus 0.4em\relax IEEE, 2021, pp. 13\,438--13\,444.

\bibitem{breyer2021volumetric}
M.~Breyer, J.~J. Chung, L.~Ott, R.~Siegwart, and J.~Nieto, ``Volumetric grasping network: Real-time 6 dof grasp detection in clutter,'' in \emph{Conference on Robot Learning}.\hskip 1em plus 0.5em minus 0.4em\relax PMLR, 2021, pp. 1602--1611.

\bibitem{zheng2022vgpn}
L.~Zheng, Y.~Cai, T.~Lu, and S.~Wang, ``Vgpn: 6-dof grasp pose detection network based on hough voting,'' in \emph{2022 IEEE/RSJ International Conference on Intelligent Robots and Systems (IROS)}.\hskip 1em plus 0.5em minus 0.4em\relax IEEE, 2022, pp. 7460--7467.

\bibitem{zheng2023gpdan}
L.~Zheng, W.~Ma, Y.~Cai, T.~Lu, and S.~Wang, ``Gpdan: Grasp pose domain adaptation network for sim-to-real 6-dof object grasping,'' \emph{IEEE Robotics and Automation Letters}, vol.~8, no.~8, pp. 4585--4592, 2023.

\bibitem{fang2023anygrasp}
H.-S. Fang, C.~Wang, H.~Fang, M.~Gou, J.~Liu, H.~Yan, W.~Liu, Y.~Xie, and C.~Lu, ``Anygrasp: Robust and efficient grasp perception in spatial and temporal domains,'' \emph{IEEE Transactions on Robotics}, vol.~39, no.~5, pp. 3929--3945, 2023.

\bibitem{mildenhall2020nerf}
B.~Mildenhall, P.~P. Srinivasan, M.~Tancik, J.~T. Barron, R.~Ramamoorthi, and R.~Ng, ``Nerf: Representing scenes as neural radiance fields for view synthesis,'' in \emph{European Conference on Computer Vision}.\hskip 1em plus 0.5em minus 0.4em\relax Springer, 2020, pp. 405--421.

\bibitem{ichnowski2022dex}
J.~Ichnowski, Y.~Avigal, J.~Kerr, and K.~Goldberg, ``Dex-nerf: Using a neural radiance field to grasp transparent objects,'' in \emph{Conference on Robot Learning}.\hskip 1em plus 0.5em minus 0.4em\relax PMLR, 2022, pp. 526--536.

\bibitem{dai2023graspnerf}
Q.~Dai, Y.~Zhu, Y.~Geng, C.~Ruan, J.~Zhang, and H.~Wang, ``Graspnerf: Multiview-based 6-dof grasp detection for transparent and specular objects using generalizable nerf,'' in \emph{2023 IEEE International Conference on Robotics and Automation (ICRA)}.\hskip 1em plus 0.5em minus 0.4em\relax IEEE, 2023, pp. 1757--1763.

\bibitem{liu2024rgbgrasp}
C.~Liu, K.~Shi, K.~Zhou, H.~Wang, J.~Zhang, and H.~Dong, ``Rgbgrasp: Image-based object grasping by capturing multiple views during robot arm movement with neural radiance fields,'' \emph{IEEE Robotics and Automation Letters}, vol.~9, no.~6, pp. 6012--6019, 2024.

\bibitem{tancik2023nerfstudio}
M.~Tancik, E.~Weber, E.~Ng, R.~Li, B.~Yi, T.~Wang, A.~Kristoffersen, J.~Austin, K.~Salahi, A.~Ahuja \emph{et~al.}, ``Nerfstudio: A modular framework for neural radiance field development,'' in \emph{ACM SIGGRAPH 2023 Conference Proceedings}, 2023, pp. 1--12.

\bibitem{barron2021mip}
J.~T. Barron, B.~Mildenhall, M.~Tancik, P.~Hedman, R.~Martin-Brualla, and P.~P. Srinivasan, ``Mip-nerf: A multiscale representation for anti-aliasing neural radiance fields,'' in \emph{2021 IEEE/CVF International Conference on Computer Vision (ICCV)}, 2021, pp. 5835--5844.

\bibitem{barron2022mip}
J.~T. Barron, B.~Mildenhall, D.~Verbin, P.~P. Srinivasan, and P.~Hedman, ``Mip-nerf 360: Unbounded anti-aliased neural radiance fields,'' in \emph{2022 IEEE/CVF Conference on Computer Vision and Pattern Recognition (CVPR)}, 2022, pp. 5460--5469.

\bibitem{muller2022instant}
T.~M{\"u}ller, A.~Evans, C.~Schied, and A.~Keller, ``Instant neural graphics primitives with a multiresolution hash encoding,'' \emph{ACM transactions on graphics (TOG)}, vol.~41, no.~4, pp. 1--15, 2022.

\bibitem{wang2023sparsenerf}
G.~Wang, Z.~Chen, C.~C. Loy, and Z.~Liu, ``Sparsenerf: Distilling depth ranking for few-shot novel view synthesis,'' in \emph{2023 IEEE/CVF International Conference on Computer Vision (ICCV)}, 2023, pp. 9031--9042.

\bibitem{wang2021ibrnet}
Q.~Wang, Z.~Wang, K.~Genova, P.~Srinivasan, H.~Zhou, J.~T. Barron, R.~Martin-Brualla, N.~Snavely, and T.~Funkhouser, ``Ibrnet: Learning multi-view image-based rendering,'' in \emph{2021 IEEE/CVF Conference on Computer Vision and Pattern Recognition (CVPR)}, 2021, pp. 4688--4697.

\bibitem{liu2022neural}
Y.~Liu, S.~Peng, L.~Liu, Q.~Wang, P.~Wang, C.~Theobalt, X.~Zhou, and W.~Wang, ``Neural rays for occlusion-aware image-based rendering,'' in \emph{2022 IEEE/CVF Conference on Computer Vision and Pattern Recognition (CVPR)}, 2022, pp. 7814--7823.

\bibitem{yang2023contranerf}
H.~Yang, L.~Hong, A.~Li, T.~Hu, Z.~Li, G.~H. Lee, and L.~Wang, ``Contranerf: Generalizable neural radiance fields for synthetic-to-real novel view synthesis via contrastive learning,'' in \emph{2023 IEEE/CVF Conference on Computer Vision and Pattern Recognition (CVPR)}, 2023, pp. 16\,508--16\,517.

\bibitem{wang2021neus}
P.~Wang, L.~Liu, Y.~Liu, C.~Theobalt, T.~Komura, and W.~Wang, ``Neus: Learning neural implicit surfaces by volume rendering for multi-view reconstruction,'' \emph{Advances in Neural Information Processing Systems}, vol.~34, pp. 27\,171--27\,183, 2021.

\bibitem{long2022sparseneus}
X.~Long, C.~Lin, P.~Wang, T.~Komura, and W.~Wang, ``Sparseneus: Fast generalizable neural surface reconstruction from sparse views,'' in \emph{European Conference on Computer Vision}.\hskip 1em plus 0.5em minus 0.4em\relax Springer, 2022, pp. 210--227.

\bibitem{xu2023c2f2neus}
L.~Xu, T.~Guan, Y.~Wang, W.~Liu, Z.~Zeng, J.~Wang, and W.~Yang, ``C2f2neus: Cascade cost frustum fusion for high fidelity and generalizable neural surface reconstruction,'' in \emph{2023 IEEE/CVF International Conference on Computer Vision (ICCV)}, 2023, pp. 18\,245--18\,255.

\bibitem{ren2023volrecon}
Y.~Ren, F.~Wang, T.~Zhang, M.~Pollefeys, and S.~Süsstrunk, ``Volrecon: Volume rendering of signed ray distance functions for generalizable multi-view reconstruction,'' in \emph{2023 IEEE/CVF Conference on Computer Vision and Pattern Recognition (CVPR)}, 2023, pp. 16\,685--16\,695.

\bibitem{liang2024retr}
Y.~Liang, H.~He, and Y.~Chen, ``Retr: Modeling rendering via transformer for generalizable neural surface reconstruction,'' \emph{Advances in Neural Information Processing Systems}, vol.~36, 2024.

\bibitem{na2024uforecon}
Y.~Na, W.~J. Kim, K.~B. Han, S.~Ha, and S.-E. Yoon, ``Uforecon: Generalizable sparse-view surface reconstruction from arbitrary and unfavorable sets,'' in \emph{Proceedings of the IEEE/CVF Conference on Computer Vision and Pattern Recognition}, 2024, pp. 5094--5104.

\bibitem{kerr2023evo}
J.~Kerr, L.~Fu, H.~Huang, Y.~Avigal, M.~Tancik, J.~Ichnowski, A.~Kanazawa, and K.~Goldberg, ``Evo-nerf: Evolving nerf for sequential robot grasping of transparent objects,'' in \emph{Conference on Robot Learning}.\hskip 1em plus 0.5em minus 0.4em\relax PMLR, 2023, pp. 353--367.

\bibitem{lin2023mira}
Y.-C. Lin, P.~Florence, A.~Zeng, J.~T. Barron, Y.~Du, W.-C. Ma, A.~Simeonov, A.~R. Garcia, and P.~Isola, ``Mira: Mental imagery for robotic affordances,'' in \emph{Conference on Robot Learning}.\hskip 1em plus 0.5em minus 0.4em\relax PMLR, 2023, pp. 1916--1927.

\bibitem{sun2021loftr}
J.~Sun, Z.~Shen, Y.~Wang, H.~Bao, and X.~Zhou, ``Loftr: Detector-free local feature matching with transformers,'' in \emph{2021 IEEE/CVF Conference on Computer Vision and Pattern Recognition (CVPR)}, 2021, pp. 8918--8927.

\bibitem{piccardi2004background}
M.~Piccardi, ``Background subtraction techniques: a review,'' in \emph{2004 IEEE International Conference on Systems, Man and Cybernetics (IEEE Cat. No.04CH37583)}, vol.~4, 2004, pp. 3099--3104 vol.4.

\bibitem{ronneberger2015u}
O.~Ronneberger, P.~Fischer, and T.~Brox, ``U-net: Convolutional networks for biomedical image segmentation,'' in \emph{Medical image computing and computer-assisted intervention--MICCAI 2015: 18th international conference, Munich, Germany, October 5-9, 2015, proceedings, part III 18}.\hskip 1em plus 0.5em minus 0.4em\relax Springer, 2015, pp. 234--241.

\bibitem{zhao2021point}
H.~Zhao, L.~Jiang, J.~Jia, P.~Torr, and V.~Koltun, ``Point transformer,'' in \emph{2021 IEEE/CVF International Conference on Computer Vision (ICCV)}, 2021, pp. 16\,239--16\,248.

\bibitem{coumans2016pybullet}
E.~Coumans and Y.~Bai, ``Pybullet, a python module for physics simulation for games, robotics and machine learning,'' 2016--2021, [Online]. Available: \url{http://pybullet.org}.

\bibitem{blender}
``Blender,'' [Online]. Available: \url{https://www.blender.org/}.

\bibitem{dai2022domain}
Q.~Dai, J.~Zhang, Q.~Li, T.~Wu, H.~Dong, Z.~Liu, P.~Tan, and H.~Wang, ``Domain randomization-enhanced depth simulation and restoration for perceiving and grasping specular and transparent objects,'' in \emph{European Conference on Computer Vision}.\hskip 1em plus 0.5em minus 0.4em\relax Springer, 2022, pp. 374--391.

\bibitem{zhou2022devnet}
K.~Zhou, L.~Hong, C.~Chen, H.~Xu, C.~Ye, Q.~Hu, and Z.~Li, ``Devnet: Self-supervised monocular depth learning via density volume construction,'' in \emph{European Conference on Computer Vision}.\hskip 1em plus 0.5em minus 0.4em\relax Springer, 2022, pp. 125--142.

\bibitem{van2008visualizing}
L.~Van~der Maaten and G.~Hinton, ``Visualizing data using t-sne.'' \emph{Journal of machine learning research}, vol.~9, no.~11, 2008.

\end{thebibliography}

\end{document}